\documentclass[conference]{IEEEtran}
\IEEEoverridecommandlockouts
\usepackage{cite}
\usepackage{amsmath,amssymb,amsfonts}
\usepackage{algorithmic}
\usepackage[ruled]{algorithm2e}
\usepackage{graphicx}
\usepackage{textcomp}
\usepackage{xcolor}
\usepackage{pifont}
\usepackage[colorlinks]{hyperref}
\usepackage{booktabs}
\usepackage{multirow}
\usepackage{marvosym}
\def\BibTeX{{\rm B\kern-.05em{\sc i\kern-.025em b}\kern-.08em
    T\kern-.1667em\lower.7ex\hbox{E}\kern-.125emX}}

\makeatletter
\newcommand{\linebreakand}{%
  \end{@IEEEauthorhalign}
  \hfill\mbox{}\par
  \mbox{}\hfill\begin{@IEEEauthorhalign}
}
\makeatother

\begin{document}

\title{FedStyle: Style-Based Federated Learning Crowdsourcing Framework for Art Commissions\\
\thanks{This work is supported by The National Natural Science Foundation of China (No. 62172155). Corresponding authors: Yeting Guo and Fang Liu.} 
}

\author{\IEEEauthorblockN{1\textsuperscript{st} Changjuan Ran}
\IEEEauthorblockA{\textit{School of Design} \\
\textit{Hunan University}\\
Changsha, China \\
s220800533@hnu.edu.cn}
\and
\IEEEauthorblockN{2\textsuperscript{nd} Yeting Guo\textsuperscript{\Letter}}
\IEEEauthorblockA{\textit{College of Computer} \\
\textit{National University of Defense Technology}\\
Changsha, China \\
guoyeting13@nudt.edu.cn}
\and
\IEEEauthorblockN{3\textsuperscript{rd} Fang Liu\textsuperscript{\Letter}}
\IEEEauthorblockA{\textit{School of Design} \\
\textit{Hunan University}\\
Changsha, China \\
fangl@hnu.edu.cn} 
\linebreakand
\IEEEauthorblockN{4\textsuperscript{th} Shenglan Cui}
\IEEEauthorblockA{\textit{School of Design} \\
\textit{Hunan University}\\
Changsha, China \\
cuisl@hnu.edu.cn}
\and
\IEEEauthorblockN{5\textsuperscript{th} Yunfan Ye}
\IEEEauthorblockA{\textit{College of Computer} \\
\textit{National University of Defense Technology}\\
Changsha, China \\
yunfan951202@gmail.com}

}

\maketitle

\begin{abstract}
The unique artistic style is crucial to artists' occupational competitiveness, yet prevailing Art Commission Platforms rarely support style-based retrieval. Meanwhile, the fast-growing generative AI techniques aggravate artists' concerns about releasing personal artworks to public platforms. To achieve artistic style-based retrieval without exposing personal artworks, we propose FedStyle, a style-based federated learning crowdsourcing framework. It allows artists to train local style models and share model parameters rather than artworks for collaboration. However, most artists possess a unique artistic style, resulting in severe model drift among them. FedStyle addresses such extreme data heterogeneity by having artists learn their abstract style representations and align with the server, rather than merely aggregating model parameters lacking semantics. Besides, we introduce contrastive learning to meticulously construct the style representation space, pulling artworks with similar styles closer and keeping different ones apart in the embedding space. Extensive experiments on the proposed datasets demonstrate the superiority of FedStyle.
\end{abstract}


\section{Introduction}
Numerous independent artists in the world are registered in art commission platforms to find possible commercial orders~\cite{glaze}.
These platforms enable artists to exhibit artwork examples, await requests, or find user-posted solicitations, such as Artistree, Fiverr, MadeMay, etc. 
Individual artistic styles serve as distinctive markers to identify artists, reflecting the years of training professional artists undergo to cultivate their unique artistic ideals~\cite{glaze, Philosophical-personal, hopkins2021artistic}. 
However, users seeking commissions, often lacking professional art expertise, struggle to articulate specific individual artistic styles or recall artists aligned with such styles.At best, they can find similar style reference images. Consequently, they are compelled to engage in ceaseless searches or adopt a passive stance, awaiting artists undertaking commissions, as prevailing art commission platforms mainly support keyword retrieval, leaving the style-guided artist retrieval problem unresolved.

To address the aforementioned problem, content-based image retrieval could serve as an alternative solution. However, well-established image search engines such as Google Lens, Tineye, and Yandex mainly retrieve images by global features that highlight similar salient objects, and do not support image retrieval based on similarity in artistic style, which limits their applicability in art commissions.Noticeably, an individual style is typically expressed across the artist's range of works~\cite{hopkins2021artistic}. Therefore, style classification is preferable to image retrieval in individual artistic style representation. Specifically, a style-based image classification solution with artists' names as labels is proposed for style-guided artist retrieval. To train a style classification model with satisfactory accuracy, rich artistic images from various contemporary artists should be collected. However, there are persistent challenges in collecting such datasets. First, uploading artistic images to a centralized server poses a risk of data leakage. Second, the development of AI-generated art and associated fine-tuning techniques has fueled artists' reluctance to share their works online~\cite{MIT22,ANDY22}. Even when some artists do share their artworks, they may also be watermarked~\cite{luo2023irwart} or pixel-level disturbed~\cite{glaze,liang2023mist} and unnoticeable to the human eye, making style feature extraction difficult.

In this paper, we propose \textbf{FedStyle}, a \textbf{Fed}erated Learning crowdsourcing framework for individual artistic \textbf{Style} classification. Federated learning (FL)~\cite{ICME23Wu,ICME21Fedns} is a privacy-preserving distributed learning paradigm. Motivated by FL, the artists train their own style classification model with local personal artworks and share their model parameters instead of artworks with the server for aggregating a global style classification model. Such collaborative training is not as straightforward as it seems to be in practice. Since each artist only has his/her own style of images, the locally trained model can be easily overfitting and lose the ability to distinguish the artist’s style from others. Meanwhile, the convergence direction of the local models among artists is greatly different, resulting in the poor performance of the aggregated global model. In view of the extreme data heterogeneity problem, FedStyle abstracts and learns the style representation from each artist to regularize the model training and aggregation. It consists of local style training and global style aggregation. The former systematically learns the abstract local style representation, incorporating awareness of local style prediction loss, representation disparities across various styles, and distinctions in representation relative to the server. The latter aggregates these distributed local style representations into the global model with a tiny public dataset, enabling the global model to distinguish various styles. 

The main contributions are summarized as follows:

\begin{itemize}
    \item We propose FedStyle, the first-of-its-kind style-based FL crowdsourcing framework tailored for art commissions, in which data are extremely heterogeneous.
    \item Fedstyle learns the style representations along with the model parameters to reduce the impact of data heterogenity on model performance. Besides, we design training loss by introducing contrastive learning to meticuloursly construct the style representation space.
    \item We manually collect artists' artwork based on their styles as experimental datasets. The quantitative and qualitative evaluation results showed the superiority and reasonableness of FedStyle with higher overall accuracy and F1-score compared to baselines, with the satisfaction from both buyers and artists to FedStyle.
\end{itemize}

\section{Related Work}
\noindent\textbf{Style Representation and Learning.}
Early efforts by Karayev et al.~\cite{14BMVC} laid the groundwork for artistic style classification by utilizing layer activations from a Convolutional Neural Network, followed by numerous similar studies~\cite{mao2017deepart,2018imageStyleclassify,2019multitask}.
Recently, ALADIN~\cite{ALADIN_2021_ICCV} was proposed to learns fine-grained style embeddings through a multiple-stage encoder, utilizing Adaptive Instance Normalization values and a concatenated AdaIN feature set. A primary limitation of these methods is their assumption of a pre-collected centralized dataset, neglecting considerations of data privacy. Therefore, FL is introduced in the design of FedStyle.

\noindent\textbf{Federated Learning.}
As a privacy-preserving distributed learning paradigm, FL is widely used in healthcare, finance, advertising, and various domains~\cite{scaffold}. However, the local data among clients in FL are usually heterogeneous, making the global aggregated model hard to converge. Many researchers are devoted to solving this problem. FedProx~\cite{li2020fedprox} incorporates a proximal term to minimize client drift by preventing local parameters from being too far from the global parameters after updates. SCAFFOLD~\cite{scaffold} and FedDC~\cite{gao2022feddc} use gradient calibration to fix local drift. Additionally, studies such as MOON~\cite{MOON}, FedProto~\cite{tan2022fedproto}, and FedProc~\cite{mu2023fedproc} try to align the local model with the global model at the feature level through local loss design. A recent work, FPL~\cite{huang2023FPL}, focuses on domain shift and introduces a consistency regularization to better cluster prototypes and construct unbiased prototypes. Most of the aforementioned methods assume that each client has two or more classes of data, failing to cover the case that there is no overlap in data classes between clients, which exists in art commissions.

\begin{figure}[ht]
  \centering
  \includegraphics[width=1\linewidth]{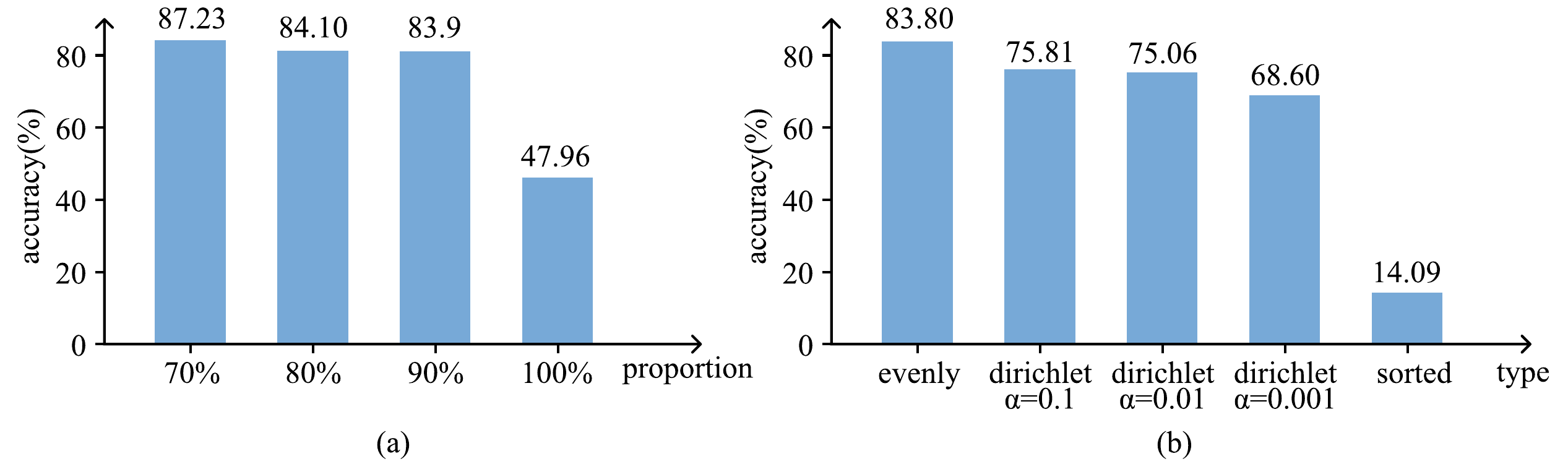}
  \caption{Accuracy under different proportion of cloaked images and different heterogeneity settings.}
  \label{fig:fig1}
\end{figure}

\section{Motivation}
As artists may use watermarking to prevent personal artworks from being used for model training in the context of generative AI, we have conducted 2 pre-experiments, trying to figure out 2 questions: whether the presence of cloaked images in the centralized dataset affects the performance of style classification model; how well the classical FedAvg~\cite{mcmahan2017communication} works on heterogeneous data distribution settings. We train Artiststyle dataset (cf. \ref{ExperimentSetup}) using Resnet-18~\cite{he2016deep}.

\noindent\textbf{Poor style classification under the cloaked images.} Fig. \ref{fig:fig1}(a) illustrates the style classification performance when cloaked images occupy different proportions of the dataset. We randomly select 200 images of 5 classes from Artiststyle for training and cloak some of them by Mist~\cite{liang2023mist}, which is a promising image watermarking tool designed for image style protection. As shown in Fig. \ref{fig:fig1}(a), the inclusion of cloaked images detracts the model from performing learning task, for the test accuracy declining with the proportion increasing. With the increasing concerns on the data privacy, artists tend to provide cloaked images, but they could not support to train a high-performance model. Thus, we are motivated to provide a distributed crowdsourcing learning solution, which supports artists to train an individual artistic style classification model collaboratively without sharing their artworks.

\noindent\textbf{Poor style classification under the highly heterogeneous images}. Fig. \ref{fig:fig1}(b) shows the performance of the global model under different data heterogeneity simulations using FedAvg algorithm. \ding{202} \emph{Evenly} means that the data is evenly and randomly distributed among clients. \ding{203} \emph{Dirichlet distribution Dir($\alpha$)} is used to simulate the heterogeneous settings, with a smaller $\alpha$ indicating the higher heterogeneity. \ding{204} \emph{Sorted} means each client has only one class of data from the whole dataset, representing the extreme data heterogeneity. It's observed that the accuracy reaches the lowest value when data are arranged following sorted distribution, compared to the highest value when data are distributed evenly. Obviously, the highly heterogeneous setting leads to the serious model drift and significant negative impact on the aggregated global model. It motivates us to explore solutions to address the extreme data heterogeneity problem in FL.

\begin{figure}[htbp]
  \centering
  \includegraphics[width=1\linewidth]{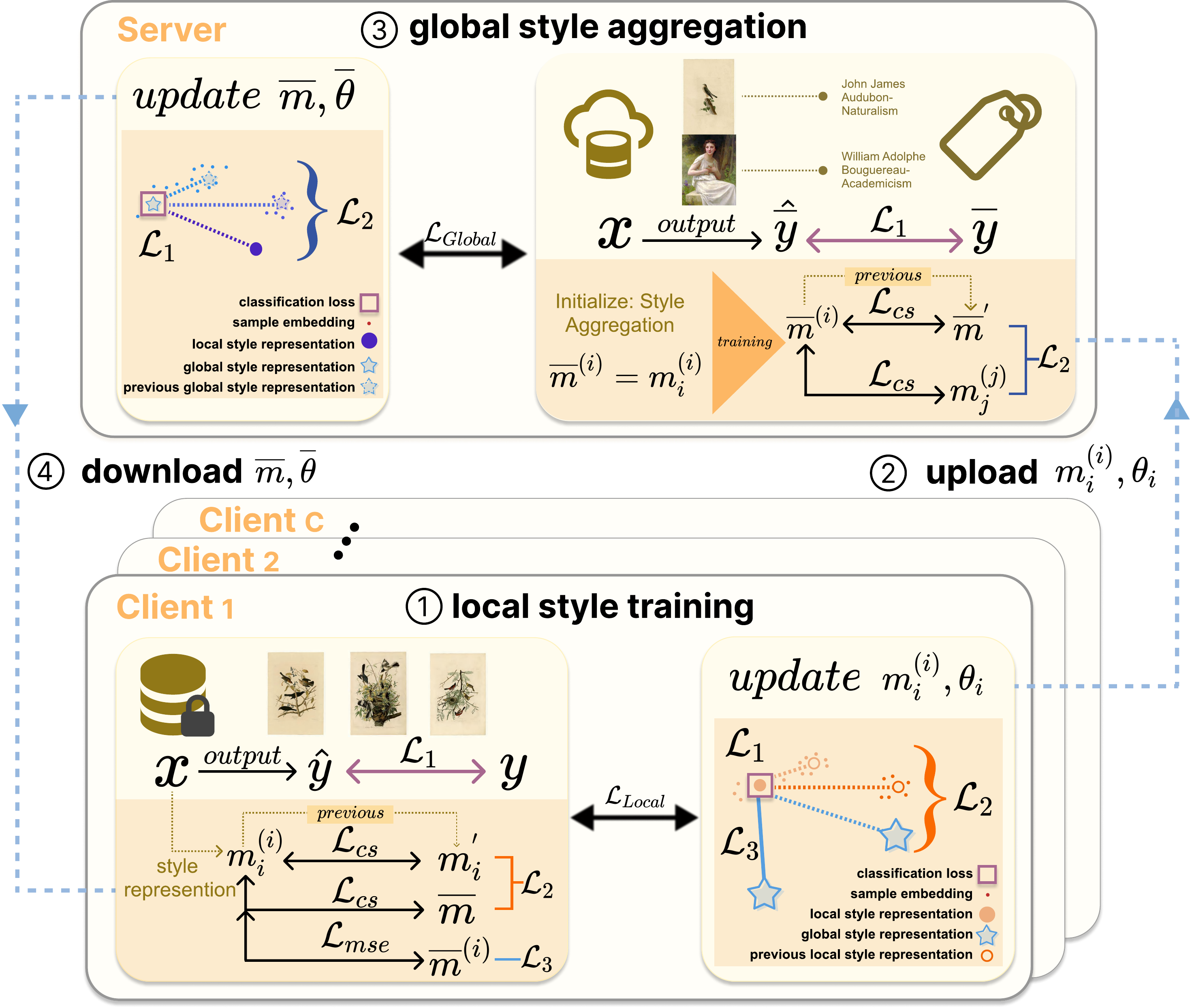}
  \caption{System framework overiew.}
  \label{fig:framwork}
\end{figure}

\section{Methodology}
Facing the above challenges, we propose FedStyle that learns the style representation for each artist and aggregates all the representations into the global model.

\subsection{System Framework}
Fig.\ref{fig:framwork} shows the overview of FedStyle. It is essentially a iterative optimization loop of the following four steps.

\ding{202} Each client trains its local model and calculates the representation of local unique style based on its personal artworks. During the training, the client would take into account the the local style prediction loss, representation differences with different styles and representation differences with the server. In that way, it reduces the model drift and increases the style distinguishability among clients. 

\ding{203} Then they upload their local style representations and model parameters to the central server. 

\ding{204} The server weights these distributed local models to form a global model, and further performs global model training to aggregate the distributed local style representations. In that way, the global model obtains knowledge about various style representations and reduces the negative impact of model drift on the global model convergence. 

\ding{205} Finally, the global model parameters and style representations are sent back to the client for further optimization. 

Next we would describe the two key steps of local style learning and global style aggregation in detail.

\subsection{Local Style Training}
To get a representative abstract style representation, we extract the average embedding vector of the penultimate network layer as the local style representation, and then optimize the representation by keeping it away from other style representations and keeping close with the corresponding global ones and local historical ones. 

Given images of style $i$ from the $i$-th client's personal artworks $D_i$, the style is the average value of the embedding vectors of these images,
\begin{equation}
\setlength{\abovedisplayskip}{0.2cm}
\setlength{\belowdisplayskip}{0cm}
m_{i}^{(i)} = \frac{1}{\left |D_{i} \right | } \sum_{(x,y)\in D_{i}}{f(x;\theta_i)} 
\end{equation}
\noindent where $m_i^{(i)}$ denotes the style representation of client $i$, with the subscript denoting the index of the local client and superscript denoting the class index. $x$ is the input image and $y$ is the label. $f(\theta_i)$ is the embedding function of client $i$, parameterized by $\theta_i$, and $f(x;\theta_i)$ denotes the embeddings of $x$, extracted from the penultimate layer of a classification model.

Considering local personal artworks in each client is a unique class of data, the local classification model is prone to overfitting. Therefore, to effectively leverage label information, we propose to use contrastive learning~\cite{khosla2020supervised} for local model training, in addition to the generally used loss of classification error. The loss function consists of 3 parts. \ding{202} $\mathcal{L}_1$ minimizes the classification error. \ding{203} $\mathcal{L}_2$ tries to pull the sample embedding close to the style they belong to and keep them isolated from other styles simultaneously. \ding{204} $\mathcal{L}_3$ is designed to align the local style with the respective unbiased global style. The local loss function is defined as follows: 
\begin{equation}
\setlength{\abovedisplayskip}{0.2cm}
\setlength{\belowdisplayskip}{0cm}
\begin{aligned} 
\mathcal{L}_L(D_i)&=\lambda_1\cdot \mathcal{L}_1(\widehat{y},y)+\lambda_2\cdot \mathcal{L}_2(m_i^{(i)},m_i^{'},\overline{m})\\
&+\lambda_3\cdot \mathcal{L}_3(m_i^{(i)},\overline{m})\\
\label{local_loss}
\end{aligned}
\end{equation}
where $\lambda_1$, $\lambda_2$ and $\lambda_3$ are importance weights. $\mathcal{L}_1$ is the loss of classification error, $\mathcal{L}_2$ is the contrastive loss and $\mathcal{L}_3$ calculates the mean squared error (MSE) loss. $y$ is the label of $x$, and $\widehat{y}$ is the output. $m_i^{'}$ denotes the style sets of client $i$ in the previous round, and $\overline{m}$ is the global styles, with the overlining denotes that the parameter is from the server. $\mathcal{L}_2$ is defined as:
\begin{equation}
\setlength{\abovedisplayskip}{0.2cm}
\setlength{\belowdisplayskip}{0cm}
\begin{aligned} 
\mathcal{L}_2(m_i^{(i)},m_i^{'},\overline{m})&={\sum_{j=1}^{C}(\mathcal{L}_{cs}(m_i^{(i)},m_i^{{(j)}'})+\mathcal{L}_{cs}(m_i^{(i)},\overline{m}^{(j)}))}\\
&/(\mathcal{L}_{cs}(m_i^{(i)},m_i^{{(i)}'})+\mathcal{L}_{cs}(m_i^{(i)},\overline{m}^{(i)}))
\end{aligned}
\end{equation}
where $\mathcal{L}_{cs}$ calculates the exponential value of the cosine similarity. $\mathcal{L}_3$ is defined as:
\begin{equation}
\mathcal{L}_3(m_i^{(i)},\overline{m})=\mathcal{L}_{mse}(m_i^{(i)},\overline{m}^{(i)})
\end{equation}
The procedure of local style training is shown is Algorithm 1.
\begin{algorithm}
\caption{Local style training}\label{local_algorithm}
\SetKwInOut{Input}{Input}
\SetKw{Return}{return}
\For{each local epoch}{
    \For{batch $(x_i,y_i)\in D_i$}{
        Compute local style $m_i^{(i)}$ by Eq.1\;
        Compute $\mathcal{L}_L(D_i)$ by Eq.2 using local styles\;
        Update local model according to $\mathcal{L}_L(D_i)$\;
    }
}
\end{algorithm}
\subsection{Global Style Aggregation}
The central server is dedicated to training a global classification model from distributed individual artistic styles. Specifically, we collect some public artistic images from the Internet for global model training. The optimization objective for the server is to cluster samples of the same artistic style and furthermore separate different styles by contrasting with local styles. The global loss function can be formulated as
\begin{equation}
\begin{aligned} 
\mathcal{L}_G(\overline{D}^{(i)} )&=\lambda_4\cdot \mathcal{L}_1(\widehat{\overline{y}},\overline{y})+\lambda_5\cdot \mathcal{L}_2(\overline{m}^{(i)},\overline{m}^{'},m_j^{(j)}),\\
&j\in\{1,2,\cdots,C\}\\
\end{aligned}
\end{equation}
where $\lambda_4$ and $\lambda_5$ is importance weights. $\mathcal{L}_1$ is the loss of classification and $\mathcal{L}_2$ is the contrastive loss. $\widehat{\overline{y}}$ is the label of $x$ from style $i$ in the public dataset $\overline{D}^{(i)}$, and $\overline{y}$ is the output. $\overline{m}^{(i)}$ and $\overline{m}^{'}$ denote the global style $i$ and the previous round global style sets respectively. 
$m_j^{(j)}$ denotes the collected local style. $C$ is the total number of categories, equal to the total number of local clients. $\mathcal{L}_2$ is defined as:
\begin{equation}
\begin{aligned}
\mathcal{L}_2(\overline{m}^{(i)},\overline{m}^{'},m_j^{(j)})&={\sum_{j=1}^{C}(\mathcal{L}_{cs}(\overline{m}^{(i)},m_j^{{(j)}})+\mathcal{L}_{cs}(\overline{m}^{(i)},\overline{m}^{{(j)}'}))}\\
&/(\mathcal{L}_{cs}(\overline{m}^{(i)},m_i^{(i)})+\mathcal{L}_{cs}(\overline{m}^{(i)},\overline{m}^{{(i)}'}))
\end{aligned}
\end{equation}

The procedure of global style aggregation is shown in Algorithm 2. 
\begin{algorithm}
\caption{Global style aggregation}\label{global_algorithm}
\SetKwInOut{Input}{input}
\SetKw{Return}{return}
Initialize global style set $\{\overline{m}^{(j)}\}$\;
\For{each round T = 1,2, ...}{
    \For{each client $i$}{
        $\overline{m}^{(i)}\leftarrow$ LocalUpdate $(i, \overline{m},\overline{\theta})$\;
    }
    Compute $\mathcal{L}_G(\overline{D}^{(i)} )$ by Eq.5 using global styles\;
    Update global model according to $\mathcal{L}_G(\overline{D}^{(i)} )$\;
    Update global style set $\overline{m}$\;
}
\end{algorithm}

\section{Experiment and Evaluation}
\subsection{Experiment Setup}\label{ExperimentSetup}
\noindent\textbf{Datasets.} Different from the general style mentioned in public artwork datasets like Wikiart~\cite{saleh2015large}, the individual artistic style represents the distinctive essence of an artist's artworks~\cite{hopkins2021artistic}. The style remains stable within a period and may change over his/her career, as also stated by many contemporary artists in a survey~\cite{glaze}, leading to the inapplicability of Wikiart sorted by artist in this evaluation. To simulate our targeted scenario as realistically as possible, we recruited 5 artists with over 10 years of experience to help collect artworks with consistent individual artistic styles. We collected two datasets, Artiststyle and Conllustration. 
\textbf{Artiststyle} comprises 5624 artwork images of 26 classes, refined from Wikiart. 
\textbf{Conllustration} is an illustration dataset containing 3238 artworks from 18 contemporary artists. \href{https://github.com/ZpFhnu/Artiststyle18}{Artiststyle} is public online and Conllustration will be made available on request for research only, considering contemporary artists' copyright. More details are in the supplements.

In quantitative experiments, 18 classes out of 26 in Artiststyle and all classes in Conllustration are used for training, respectively. The rest 8 classes of data in Artiststyle are allocated for the framework scalability experiment. Each client has one class of data. It corresponds to each artist having a unique artistic style. Each client uses 80$\%$ and 20$\%$ data for training and testing, respectively. We select 10$\%$ data from the training dataset as the public dataset in the central server. 

\noindent\textbf{Model settings and Implementation Details.} We leverage Resnet-18~\cite{he2016deep} to train the art images. It consists of 17 convolutional layers, 2 batch normalization layers, 4 MaxPool layers, and 4 residual blocks. All the experiments were performed on an NVIDIA 6000 machine, using the Pytorch framework.
We use the SGD optimizer with a learning rate of 0.01 and momentum of 0.5. The entire FL training process lasts for 100 rounds to ensure convergence. The client uploads to the server every 5 local training epochs, and the local batch size is 32 in all experiments.

\noindent\textbf{Baselines.} We compare FedStyle with the following baselines. \ding{202} \textbf{Local} refers to training models with the public dataset. \ding{203} \textbf{FedAvg}~\cite{mcmahan2017communication} is a classical FL baseline that averages the local model as the global model. \ding{204} \textbf{FedProx}~\cite{li2020fedprox} utilizes the Euclidean distance between the global model parameters and the local ones as the additional regularization term of the loss function. \ding{205} \textbf{FedProto}~\cite{tan2022fedproto} shares the prototype of each class instead of the model parameter. \ding{206}\textbf{FPL}~\cite{huang2023FPL} constructs cluster prototypes and unbiased prototypes to provide a fair convergent target.

\begin{figure*}[ht]
  \centering
  \includegraphics[width=1\linewidth]{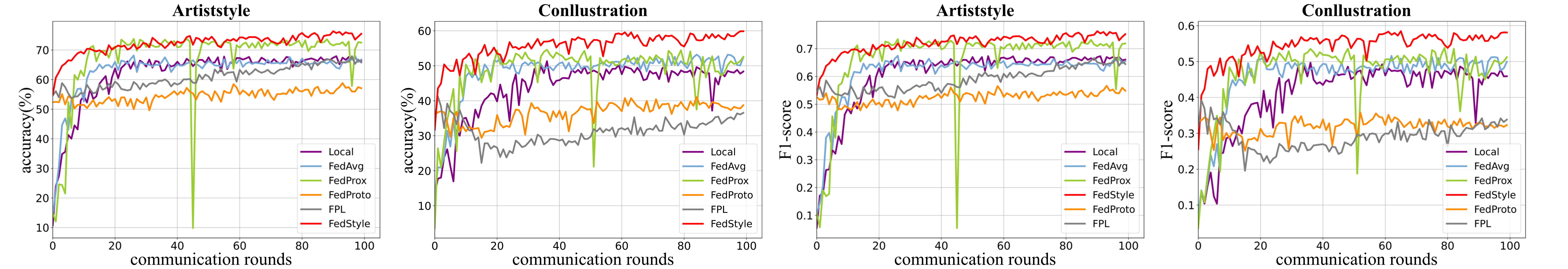}
  \caption{Comparison of accuracy and F1-score with baselines}
  \label{fig:results}
\end{figure*}

\begin{table*}[ht]
    \setlength{\tabcolsep}{4pt}
    \centering
    \tabcolsep=0.5cm
    \renewcommand\arraystretch{1.3}
    \caption{Comparison of accuracy and F1-score under different number of artists}
    \label{tab:5}  
    \begin{tabular}{l|cccccccc}
    \toprule[2pt] 
    \multirow{2}*{Method} & \multicolumn{2}{c}{N=20}  & \multicolumn{2}{c}{N=22} & \multicolumn{2}{c}{N=24} & \multicolumn{2}{c}{N=26} \\
    & Acc & F1-score & Acc & F1-score & Acc & F1-score & Acc & F1-score\\
    \cline{1-9} 
    Local & 61.96 & 60.88 & 58.13 & 57.29 & 48.42 & 47.69 & 53.51 & 59.41\\
    FedAvg & 61.86 & 60.91 & 59.75 & 57.55 & 52.30 & 51.50 & 52.36 & 51.44\\
    FedProx & 69.36 & \textbf{69.62} & 59.72 & 57.60 & 56.96 & 55.40 & 58.27 & 56.45\\
    FedProto & 55.31 & 54.10 & 57.92 & 55.43 & 51.15 & 49.90 & 49.18 & 47.78\\
    FPL & 57.93 & 56.23 & 56.96 & 55.56 & 50.95 & 50.46 & 51.58 & 49.59\\
    FedStyle & \textbf{69.68} & 69.23 & \textbf{64.78} & \textbf{63.64} & \textbf{63.53} & \textbf{62.12} & \textbf{62.22} & \textbf{60.97}\\
    \bottomrule[2pt] 
    \end{tabular}
\end{table*}

\begin{figure*}[ht]
  \centering
  \includegraphics[width=1\linewidth]{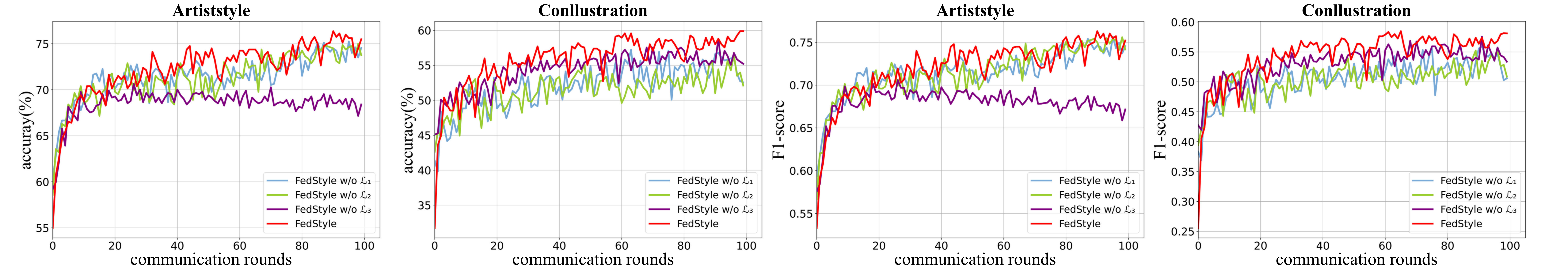}
  \caption{Ablation study results}
  \label{fig:xiaoroong}
\end{figure*}

\subsection{Quantitative Evaluation}
We evaluate the accuracy and F1-score of FedStyle and baselines on the two datasets. For Artiststyle, $\lambda_1$, $\lambda_2$, $\lambda_3$, $\lambda_4$, $\lambda_5$ are set to 10, 0.05, 20, 10, 0.005 respectively. For Conllustration, the values of $\lambda_1$ and $\lambda_2$ are set to 40 and 0.5, respectively. 

The results are as shown in Fig.\ref{fig:results}. It's observed that our FedStyle obtains optimal or suboptimal accuracy and F1-score on both the two datasets. Specifically, in the last 10 rounds of training on Artiststyle, the average accuracy of FedStyle is 75.12\%, followed by FedProx, FedAvg, Local, FPL and FedProto, with accuracy of 72.76\%, 68.07\%, 67.24\%, 63.27\% and 57.95\%, respectively. Besides, compared to FedProx, FedAvg and Local, our FedStyle achieves model convergence with fewer communication rounds and more stability. From the four line charts, it appears that the performance of FedProto and FPL has been significantly degraded, even worse than the local learning stage (Local) without federated communication, while FedStyle outperforms the best baseline (FedProx) during most of the communication rounds. Local models of each artist may converge in different directions due to extreme heterogeneity, resulting in local optima rather than global optima and degrading FL performance. The results demonstrates that our FedStyle is more effective in the collaborative artistic style learning and classification application, which exhibits the extreme data heterogeneity. 

\noindent\textbf{Framework Scalability.}
We further evaluate the model performance under different number of clients. The results are shown in Table \ref{tab:5}. The $N$ stands for the number of classes, which is also the number of clients. It clearly depicts that compared with other baselines, FedStyle consistently achieves the highest accuracy and higher F1-score in the majority of cases, verifying the scalability of our system.

\noindent\textbf{Ablation Study.} We conduct experiments to validate the effect of each loss term of designed local loss $\mathcal{L}_L$(Eq.\ref{local_loss}), shown in Fig.\ref{fig:xiaoroong}. It is found that all components help to improve the performance, showing their necessity and effectiveness. We find $\mathcal{L}_3$ has a greater impact than other components when training on Artiststyle and $\mathcal{L}_2$ a key loss term in enhancing FedStyle's performance when training on Conllustration.

\subsection{Qualitative Assessment}

\noindent\textbf{Study setup.} We invited 4 artists with over 10 years of professional painting experience and 5 buyers with demands for customized artworks to evaluate the performance of FedStyle. We randomly selected one image from each class in Artiststyle and Conllustration as query images. First, participants were asked to evaluate FedStyle's test results on query images, using a 5-point Likert scale to assess the reasonableness of the results. Second, we conducted semi-structure interviews with them after system testing.

\noindent\textbf{Results.} The statistical analysis results are shown in Table \ref{tab:4}. Avg, D-value, and S.D. represent the mean, range, and standard deviation, respectively. All average scores have passed 3 (neutral), indicating the reasonableness of the testing results given by our system. 

In the interview, artists expressed their willingness to participate in Fedstyle, noting that it can broaden the way they receive commissions while safeguarding their individual artworks. Buyers affirmed that Fedstyle can significantly save the time and effort of discovering preferred artists. All of them certified the availability of style classification. Considering the page limit, we attached more design details and results of the qualitative assessment in the supplements.

\begin{table}[htbp]
    \setlength{\tabcolsep}{4pt}
    \centering
    \tabcolsep=0.3cm
    \renewcommand\arraystretch{1.4}
    \caption{Satisfaction scale results for FedStyle}
    \label{tab:4}  
    \begin{tabular}{c|l|ccc}
    \hline
    Participation & Data Source & Avg & D-value & S.D.  \\
    \hline
    \multirow{2}{*}{Buyer} & Artiststyle & 4.52 & 3.00 & 0.79 \\ 
    \cline{2-5}
    & Conllustration & 3.82 & 3.00 & 0.85 \\
    \hline
    \multirow{2}{*}{Artist} & Artiststyle & 4.68 & 2.00 & 0.55 \\
    \cline{2-5}
    & Conllustration & 4.17 & 3.00 & 0.91 \\
    \hline
    \end{tabular}
\end{table}

\section{Conclusion}
In this paper, we introduce FedStyle, a style-based federated learning crowdsourcing framework dedicated to addressing the real needs of users involved in art commissions, in which data across artists are extremely heterogeneous. It learns the style representation of each artist in addition to model parameters to reduce the model drift among artists. 
Extensive experiments demonstrate FedStyle's effectiveness.

In the future, there will be a consideration of fine-grained factors for a more professional-style analysis. Efforts will be made to address practical challenges like communication costs when deploying FedStyle in real-world scenarios. 



\bibliographystyle{IEEEbib}
\bibliography{fedstyleRefer}

\begin{thebibliography}{10}

\bibitem{glaze}
Shawn Shan, Jenna Cryan, Emily Wenger, Haitao Zheng, Rana Hanocka, and Ben~Y Zhao,
\newblock ``Glaze: Protecting artists from style mimicry by text-to-image models,''
\newblock {\em arXiv preprint arXiv:2302.04222}, 2023.

\bibitem{Philosophical-personal}
Nick Riggle,
\newblock ``Personal style and artistic style,''
\newblock {\em The Philosophical Quarterly}, vol. 65, no. 261, pp. 711--731, 2015.

\bibitem{hopkins2021artistic}
Robert Hopkins and Nick Riggle,
\newblock ``Artistic style as the expression of ideals,''
\newblock {\em Philosophers' Imprint}, vol. 21, no. NO. 8, pp. 1--18, 2021.

\bibitem{MIT22}
Melissa Heikkiläarchive,
\newblock ``This artist is dominating ai-generated art. and he’s not happy about it.,'' Sept. 2022.

\bibitem{ANDY22}
Andy Baio,
\newblock ``Invasive diffusion: How one unwilling illustrator found herself turned into an ai model,'' Nov. 2022.

\bibitem{luo2023irwart}
Yuanjing Luo, Tongqing Zhou, Fang Liu, and Zhiping Cai,
\newblock ``Irwart: Levering watermarking performance for protecting high-quality artwork images,''
\newblock in {\em Proceedings of the ACM Web Conference 2023}, 2023, pp. 2340--2348.

\bibitem{liang2023mist}
Chumeng Liang and Xiaoyu Wu,
\newblock ``Mist: Towards improved adversarial examples for diffusion models,''
\newblock {\em arXiv preprint arXiv:2305.12683}, 2023.

\bibitem{ICME23Wu}
Chenrui Wu, Zexi Li, Fangxin Wang, and Chao Wu,
\newblock ``Learning cautiously in federated learning with noisy and heterogeneous clients,''
\newblock in {\em 2023 IEEE International Conference on Multimedia and Expo (ICME)}, 2023, pp. 660--665.

\bibitem{ICME21Fedns}
Yaoxin Zhuo and Baoxin Li,
\newblock ``Fedns: Improving federated learning for collaborative image classification on mobile clients,''
\newblock in {\em 2021 IEEE International Conference on Multimedia and Expo (ICME)}, 2021, pp. 1--6.

\bibitem{14BMVC}
Sergey Karayev, Matthew Trentacoste, Helen Han, Aseem Agarwala, Trevor Darrell, Aaron Hertzmann, and Holger Winnemoeller,
\newblock ``Recognizing image style,''
\newblock in {\em Proceedings of the British Machine Vision Conference. BMVA Press}, 2014.

\bibitem{mao2017deepart}
Hui Mao, Ming Cheung, and James She,
\newblock ``Deepart: Learning joint representations of visual arts,''
\newblock in {\em Proceedings of ACM MM}, 2017, pp. 1183--1191.

\bibitem{2018imageStyleclassify}
Wei-Ta Chu and Yi-Ling Wu,
\newblock ``Image style classification based on learnt deep correlation features,''
\newblock {\em IEEE Transactions on Multimedia}, vol. 20, no. 9, pp. 2491--2502, 2018.

\bibitem{2019multitask}
Simone Bianco, Davide Mazzini, Paolo Napoletano, and Raimondo Schettini,
\newblock ``Multitask painting categorization by deep multibranch neural network,''
\newblock {\em Expert Systems with Applications}, vol. 135, pp. 90--101, 2019.

\bibitem{ALADIN_2021_ICCV}
Dan Ruta, Saeid Motiian, Baldo Faieta, Zhe Lin, Hailin Jin, Alex Filipkowski, Andrew Gilbert, and John Collomosse,
\newblock ``Aladin: All layer adaptive instance normalization for fine-grained style similarity,''
\newblock in {\em ICCV}, October 2021, pp. 11926--11935.

\bibitem{scaffold}
Sai~Praneeth Karimireddy, Satyen Kale, Mehryar Mohri, Sashank Reddi, Sebastian Stich, and Ananda~Theertha Suresh,
\newblock ``Scaffold: Stochastic controlled averaging for federated learning,''
\newblock in {\em ICML}. PMLR, 2020, pp. 5132--5143.

\bibitem{li2020fedprox}
Tian Li, Anit~Kumar Sahu, Manzil Zaheer, Maziar Sanjabi, Ameet Talwalkar, and Virginia Smith,
\newblock ``Federated optimization in heterogeneous networks,''
\newblock {\em Proceedings of Machine learning and systems}, vol. 2, pp. 429--450, 2020.

\bibitem{gao2022feddc}
Liang Gao, Huazhu Fu, Li~Li, Yingwen Chen, Ming Xu, and Cheng-Zhong Xu,
\newblock ``Feddc: Federated learning with non-iid data via local drift decoupling and correction,''
\newblock in {\em CVPR}, 2022, pp. 10112--10121.

\bibitem{MOON}
Qinbin Li, Bingsheng He, and Dawn Song,
\newblock ``Model-contrastive federated learning,''
\newblock in {\em CVPR}, 2021, pp. 10713--10722.

\bibitem{tan2022fedproto}
Yue Tan, Guodong Long, Lu~Liu, Tianyi Zhou, Qinghua Lu, Jing Jiang, and Chengqi Zhang,
\newblock ``Fedproto: Federated prototype learning across heterogeneous clients,''
\newblock in {\em AAAI}, 2022, vol.~36, pp. 8432--8440.

\bibitem{mu2023fedproc}
Xutong Mu, Yulong Shen, Ke~Cheng, Xueli Geng, Jiaxuan Fu, Tao Zhang, and Zhiwei Zhang,
\newblock ``Fedproc: Prototypical contrastive federated learning on non-iid data,''
\newblock {\em Future Generation Computer Systems}, vol. 143, pp. 93--104, 2023.

\bibitem{huang2023FPL}
Wenke Huang, Mang Ye, Zekun Shi, He~Li, and Bo~Du,
\newblock ``Rethinking federated learning with domain shift: A prototype view,''
\newblock in {\em CVPR}. IEEE, 2023, pp. 16312--16322.

\bibitem{mcmahan2017communication}
Brendan McMahan, Eider Moore, Daniel Ramage, Seth Hampson, and Blaise~Aguera y~Arcas,
\newblock ``Communication-efficient learning of deep networks from decentralized data,''
\newblock in {\em Artificial intelligence and statistics}. PMLR, 2017, pp. 1273--1282.

\bibitem{he2016deep}
Kaiming He, Xiangyu Zhang, Shaoqing Ren, and Jian Sun,
\newblock ``Deep residual learning for image recognition,''
\newblock in {\em CVPR}, 2016, pp. 770--778.

\bibitem{khosla2020supervised}
Prannay Khosla, Piotr Teterwak, Chen Wang, Aaron Sarna, Yonglong Tian, Phillip Isola, Aaron Maschinot, Ce~Liu, and Dilip Krishnan,
\newblock ``Supervised contrastive learning,''
\newblock {\em Advances in neural information processing systems}, vol. 33, pp. 18661--18673, 2020.

\bibitem{saleh2015large}
Babak Saleh and Ahmed Elgammal,
\newblock ``Large-scale classification of fine-art paintings: Learning the right metric on the right feature,''
\newblock {\em arXiv preprint arXiv:1505.00855}, 2015.

\end{thebibliography}

\end{document}